%% file: iFairNMF.tex
\documentclass[runningheads]{llncs}
\usepackage[utf8]{inputenc} 
\usepackage[T1]{fontenc}    
\usepackage{hyperref}       
\usepackage{url}            
\usepackage{booktabs}       
\usepackage{amsfonts}       
\usepackage{nicefrac}       
\usepackage{microtype}      
\usepackage{xcolor}         

\usepackage{cite}
\usepackage{bm}
\usepackage{amsmath,amssymb,amsfonts}
\usepackage{algorithmic}
\usepackage{textcomp}
\usepackage{algorithm}
\usepackage{adjustbox, tabularx}
\usepackage{makecell, booktabs}
\usepackage{array, multirow}
\usepackage{arydshln}
\usepackage{comment}
\usepackage{svg}
\usepackage{caption}
\usepackage{subfig}
\usepackage{graphicx}
\graphicspath{{./Metadata/}}

\usepackage{eucal}
\newcommand{\RomanNumeralCaps}[1]
    {\MakeUppercase{\romannumeral #1}}
\usepackage{todonotes}
\begin{document}

\title{Towards Cohesion-Fairness Harmony: Contrastive Regularization in Individual Fair Graph Clustering
}

\titlerunning{individual Fair NMTF}

\author{Siamak Ghodsi\inst{1}\href{https://orcid.org/0000-0002-3306-4233}{\includegraphics[scale=0.009]{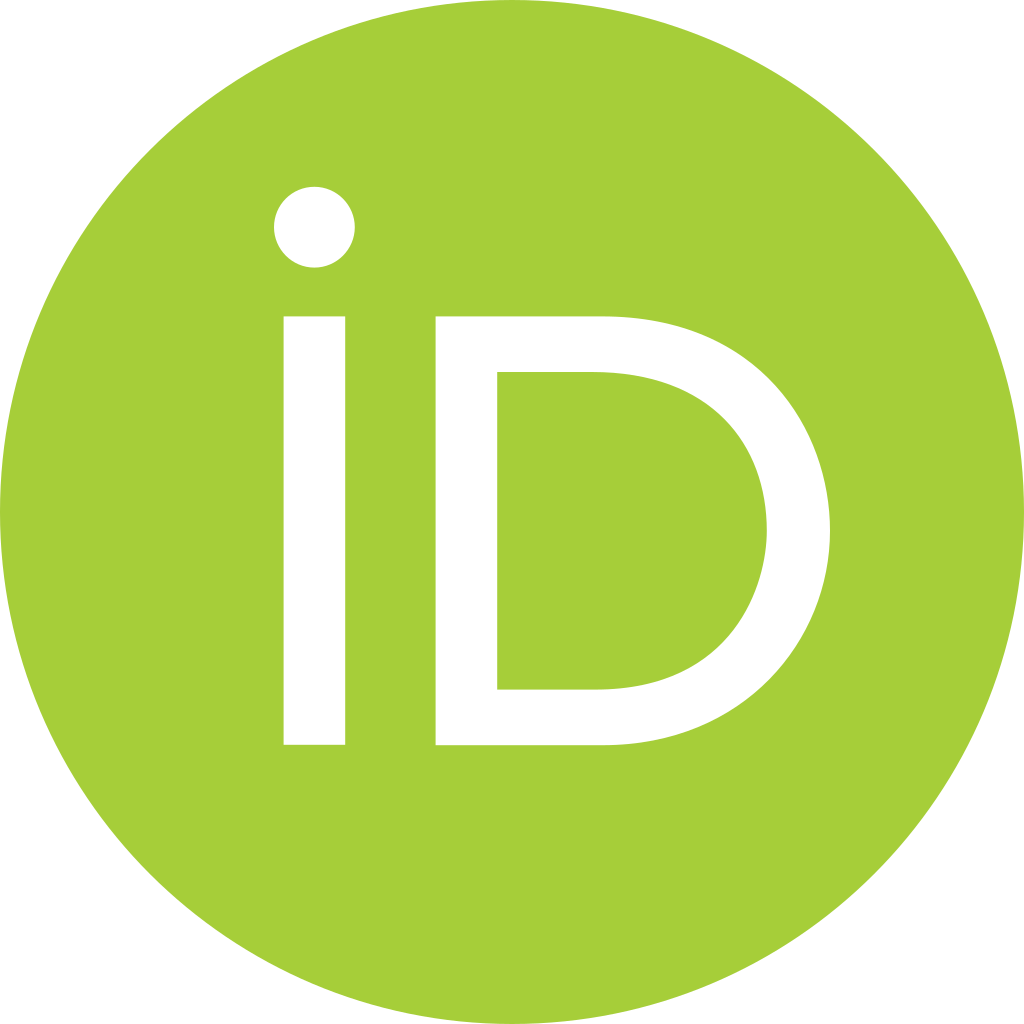}}
\and
Seyed Amjad Seyedi\inst{2}\href{https://orcid.org/0000-0003-2718-7146}{\includegraphics[scale=0.009]{orcid.png}} 
\and
Eirini Ntoutsi\inst{3}\href{https://orcid.org/0000-0001-5729-1003}{\includegraphics[scale=0.009]{orcid.png}}}
\authorrunning{S. Ghodsi et al.}
%
\institute{
L3S Research Centre, Leibniz University Hannover, Germany. \email{ghodsi@l3s.de} \and
University of Kurdistan, Sanandaj, Iran.
\email{amjadseyedi@uok.ac.ir}\and
RI CODE, University of the Bundeswehr Munich, Germany.
\scriptsize{
\email{eirini.ntoutsi@unibw.de}}
}

\maketitle 

\input{Abstract}

\section{Introduction} \label{sec: intro}
\input{Intro}

\section{Background and Related Works} \label{sec: rel_works}
\input{Related}

\section{The iFairNMTF Model} \label{sec: method}
\input{Proposed}

\section{Experimental Evaluation}\label{sec: experiments}
\input{Exp}

\section{Conclusion and Outlook} \label{sec: conclusion}
In this paper, we introduce the iFairNMTF model, an individually fair flexible approach for graph clustering that takes sensitive (node) attributes into account. iFairNMTF modifies the NMTF model's objective function by incorporating a contrastive penalty term, ensuring that clustering outcomes align with sensitive demographic information and thereby promoting individually fair cluster representations through the attraction and repulsion advantage of the proposed contrastive regularization term. The trade-off regularization parameter $\lambda$ empowers users to customize the balance between clustering performance and fairness based on their specific needs. Our experiments on both real and synthetic datasets demonstrate that adjusting the trade-off parameter allows for achieving a desired equilibrium between maximizing clustering cohesion and promoting fairness. 
Promising directions for future research include exploring multi-objective techniques to effectively balance fairness and cohesion objectives, particularly in complex, multi-dimensional discrimination scenarios~\cite{roy2023multi}. Additionally, developing NMF tailored for group fairness, with an emphasis on integrating both individual and group notions into algorithmic design. Finally, evaluating fair clustering methods, esp. for individual fairness remains an ongoing challenge. 
\input{Acknowledgement}

\bibliographystyle{splncs04}
\bibliography{References}


\end{document}

%% file: Abstract.tex
\begin{abstract}
Conventional fair graph clustering methods face two primary challenges: i) They prioritize balanced clusters at the expense of cluster cohesion by imposing rigid constraints, ii) Existing methods of both individual and group-level fairness in graph partitioning mostly rely on eigen decompositions and thus, generally lack interpretability. 
To address these issues, we propose $\emph{iFairNMTF}$, an individual Fairness Nonnegative Matrix Tri-Factorization model with contrastive fairness regularization that achieves balanced and cohesive clusters.
By introducing fairness regularization, our model allows for customizable accuracy-fairness trade-offs, thereby enhancing user autonomy without compromising the interpretability provided by nonnegative matrix tri-factorization.
Experimental evaluations on real and synthetic datasets demonstrate the superior flexibility of iFairNMTF in achieving fairness and clustering performance.

\keywords{Fair Graph Clustering \and Fair-Nonnegative Matrix Factorization \and Fair Unsupervised Learning \and Individual Fairness.}
\end{abstract}

%% file: Intro.tex
Graph-structured data is ubiquitous in various real-world applications including recommender systems, 
 e-commerce
, social networks
, and neural networks. 
Graph clustering is essential for identifying meaningful patterns within graphs.
Despite the advancements in algorithmic fairness for supervised learning scenarios which are mostly tailored for independent and identically distributed (i.i.d.) data~\cite{ntoutsi2020bias}, the topic of fairness is less explored in the unsupervised learning domain and especially for graphs. A motivating example comes from the educational domain~\cite{DBLP:conf/pakdd/QuyFN23}: how to divide students in a classroom into smaller groups for collaborative assignments. It is demanded to diversify group members from different genders or races while respecting existing friendship networks and maintaining connections.
Graphs comprise non-i.i.d. data; thus, the broad literature on fairness for i.i.d. data is generally not applicable to graphs \cite{dong2023fairness}. However, some approaches mitigate bias by converting graph data into tabular form and leveraging existing methods. Additionally, there exist bias mitigation approaches that transform tabular data into hypergraphs based on dataset similarities, e.g. \cite{DBLP:conf/ewaf/GhodsiN23}.

In the realm of fairness in i.i.d. clustering, the pioneering work of ~\cite{DBLP:conf/nips/Chierichetti0LV17} introduced balance score, a fairness measure rooted in statistical parity~\cite{DBLP:conf/innovations/DworkHPRZ12}, aiming at clusters of balanced demographic subgroups given a sensitive feature. Inspired by this work,~\cite{DBLP:conf/icml/KleindessnerSAM19} proposed a spectral graph clustering (SC) framework promoting group fairness that was later extended in~\cite{wang2023scalable} to scaled networks. However, there is not much literature on clustering with individual fairness, which prioritizes treating similar individuals (nodes in our context) similarly.
A spectral model based on PageRank was proposed in~\cite{DBLP:conf/kdd/KangHMT20} introducing a notion of individual fairness but for supervised node-classification tasks, whereas~\cite{DBLP:conf/bigdataconf/WangKXLT22} introduces an individual-fair model for multi-view graph clustering. Only in \cite{DBLP:conf/nips/GuptaD22}, an (unsupervised) graph partitioning method employed an individual fairness approach which constrains a spectral clustering with a representation graph constructed solely based on sensitive information of individuals. 
SC methods are based on minimizing either the Ratio-cut or the Normalized-cut heuristic that generally tend to minimize the number of links pointing outside each cluster \cite{DBLP:journals/sac/Luxburg07}. These cut-based heuristics do not guarantee to discover the optimal graph partitioning. Thus, incorporating (hard) fairness constraints into these rigid frameworks, which is usually also not a trivial and straightforward process, makes achieving the optimal solution challenging such that usually a relaxed form of the problem is being solved as in \cite{ DBLP:conf/icml/KleindessnerSAM19, DBLP:conf/nips/GuptaD22}. In addition, since the solution to these hard-constrained spectral approaches is based on the eigen-decomposition of the graph, it lacks interpretability.

To address the identified issues, we introduce a versatile fairness-aware model for graph clustering, the so-called individually-Fair Symmetric Nonnegative Matrix Tri-Factorization  ($\emph{iFairNMTF}$) model with contrastive regularization. Building on the symmetric NMF \cite{DBLP:conf/sdm/KuangPD12, Li2018Sep}, a model tailored for graph clustering, the NMTF \cite{DBLP:conf/ijcai/PeiCS15} extends its capabilities inheriting its intrinsic interpretability through non-negativity and direct clustering, while other models require steps like graph and/or node embedding \cite{DBLP:conf/wsdm/DaiW21, DBLP:conf/kdd/DongKTL21}, representation learning \cite{DBLP:journals/jmlr/TsitsulinPPM23}, or eigen-decomposition \cite{DBLP:conf/icml/KleindessnerSAM19, wang2023scalable, DBLP:conf/kdd/KangHMT20} before performing the final clustering. 
Additionally, NMTF provides better clustering and also introduces an explicit interpretability factor for inter-cluster interactions. We integrate these capabilities with a novel soft individual fairness regularization in
$\emph{iFairNMTF}$ 
with an adjustable parameter $\lambda$ for balancing both fairness and clustering objectives. 
Our key contributions include: i) A flexible joint learning framework with adjustable fairness regularization, accommodating customization of fairness enforcement in relation to clustering quality. The framework supports the linear integration of fairness and other problem-specific constraints via a customizable cost function. ii) Introduction of a contrastive fairness regularization, promoting the distribution of similar individuals across clusters based on sensitive attribute membership while ensuring distinct representation of dissimilar individuals within each cluster. iii) Retention of SNMF advantages, providing an interpretable data representation due to non-negativity and direct clustering. iv) Integration of an explicit interpretability factor, exposing inter-cluster relationships. v) Extensive experiments demonstrating the efficacy of our model with soft-fairness constraints and emphasizing the significance of the adjustable trade-off optimization.

To the best of our knowledge, our proposed joint learning contrastive framework is the first attempt to integrate an NMF model into a fairness-aware learning framework. The rest of this paper is organized as follows: In Section~\ref{sec: rel_works}, we review related work. Our method is introduced in Section~\ref{sec: method}. The experimental evaluation is presented in Section~\ref{sec: experiments}. Conclusions and outlook are discussed in Section~\ref{sec: conclusion}.

%% file: Related.tex
\subsubsection{Problem Formulation}
Let us assume an undirected graph $\mathcal{G} = \left(V, E\right)$ where $V = \{v_1, v_2, \ldots, v_n \}$ is the set of $n$ nodes and $E \subseteq V \times V$ is the set of edges.  The adjacency matrix $\bm{A} \in \mathbb{R}^{n \times n}$ encodes the edge information; the existence or non-existence of an edge between two nodes $v_i, v_j$ is modeled as $a_{ij} = 1$ and $a_{ij} = 0$, respectively.
Also, we assume no self-loops (edge connecting a node to itself), so $a_{ii} = 0$ for all $i \in [n]$. 
Let us further assume that the set of vertices constitutes $m$ disjoint groups identified based on a sensitive attribute e.g., gender or race, such that $V = \dot\cup_{s \in \left[m\right]} V_s$.
The goal is to find a non-overlapping clustering of $V$ into $k\geq2$ clusters $V = 
\{C_1\dot\cup \ldots \dot\cup C_k\}$ which is subject to individual fairness. 

\subsubsection{Individual Fairness}
Individual fairness primarily formalized in~\cite{DBLP:conf/innovations/DworkHPRZ12} identifies a model $f$ to be fair if, for any pair of inputs $v_i, v_j$ which are sufficiently close (as per an appropriate metric), the model outputs $f(v_i), f(v_j)$ should also be close (as per another appropriate metric). In other terms, pairwise node distances in the input space and output space should satisfy the Lipschitz continuity Condition. Specifically, it requires the distance of any node pairs in the output space to be smaller or equal to their corresponding distance in the input space (usually re-scaled by a scalar). Given a pair of nodes $v_i$ and $v_j$, the Lipschitz condition is:
\begin{equation}
    D(f(v_i), f(v_j)) \leq L \cdot d(v_i, v_j)
\end{equation}
where $f(\cdot)$ is the predictive model producing the node-level outputs (e.g., embeddings). $D(\cdot, \cdot)$ and $d(\cdot, \cdot)$ are the distance metrics of output and input space and $L$ is the Lipschitz constant that re-scales the input distance between nodes $v_i, v_j$. In order to measure individual fairness based on $L$, \cite{DBLP:conf/icml/ZemelWSPD13} proposed consistency on non-graph data with the intuition to measure the average distance of the output between each individual and its $k$-nearest neighbors such that:
\begin{equation}
    1-\frac{1}{n\cdot k}\sum^{n}_{i=1}\Big|f(x_i)-\sum_{j\in kNN(x_i)}f(x_j)\Big|
\end{equation}
where $f(x_i)$ is the probabilistic classification output for node features $x_i$ of node $v_i$ and $kNN()$ is the neighborhood of node $v_i$. In general, a larger average distance indicates a lower level of individual
fairness. 

\subsubsection{Individual Fairness for Graph Clustering}
The notion of individual fairness in graph mining~\cite{dong2023fairness} can be divided into three categories by application: i) node pair distance-based fairness, ii) node ranking-based fairness, and iii) individual fairness in graph clustering.
The core idea in the first category is the investigation of achieving individual fairness in node representation and node embedding problems based on pairwise node distances. For example, in \cite{DBLP:journals/pvldb/LahotiGW19} a notion of consistency is proposed based on a similarity matrix $S$ that characterizes node similarity in input space and can be derived from node attributes, graph topology, or domain experts. 
Moreover, in \cite{DBLP:conf/kdd/KangHMT20} a measure is proposed that calculates the similarity-weighted output discrepancy between nodes to measure unfairness. 
This metric calculates the weighted sum of pairwise node distance in the output space, where the weighting score is the pairwise node similarity. Hence for any graph mining algorithm, a smaller
value of the similarity-weighted discrepancy typically implies a
higher level of individual fairness.

The second category aims to achieve individual fairness by establishing node rankings. This involves creating two ranking lists in the input and output space, $R_1$ and $R_2$ based on a pairwise similarity matrix $S$ in the input space. The satisfaction of individual fairness is determined by the alignment of these ranking lists, ensuring that $R_1$ and $R_2$ are identical for each individual \cite{DBLP:conf/kdd/DongKTL21}.

The third category which remains relatively less explored and is the focus of our work, surveys individual-level fairness for graph clustering. 
In essence, if all neighbors of each node in a graph, are proportionally distributed to each cluster, individual fairness is then fulfilled \cite{DBLP:journals/corr/abs-2105-03714}. One of the pioneering recent works in this direction is the work of Gupta, et.al., \cite{DBLP:conf/nips/GuptaD22} according to which a clustering algorithm satisfies individual fairness for node $v_i$ if:
\begin{equation}
    \frac{|\{v_j: \bm{A}_{i,j}=1 \land v_j \in C_k\}|}{|C_k|} = \frac{|\{v_j: \bm{A}_{i,j}=1\}|}{|V|}
\end{equation}
 for all clusters $C_k$. The key intuition is that for each node, the ratio occupied by its one-hop neighbors in its cluster should be the same as the ratio occupied by its one-hop neighbors in the entire population (i.e. the main graph). 

%% file: Proposed.tex
Inspired by the individual fairness of \cite{DBLP:conf/nips/GuptaD22}, we propose a novel individual fairness regularization for graph clustering. It constitutes a contrastive graph regularization that incorporates positive and negative elements, signifying the attraction and repulsion of individuals towards their similar and dissimilar neighbors, based on a sensitive (node) attribute. By integrating this regularization into a flexible clustering framework, we introduce a unique \textbf{i}ndividually \textbf{Fair} \textbf{N}on-negative \textbf{M}atrix \textbf{T}ri-\textbf{F}actorization joint learning model (iFairNMTF).
\subsection{The iFairNMTF Model Formulation}
\label{sec:formulation}
\begin{figure}
\captionsetup{font=scriptsize} 
    \centering
    \includegraphics[width=0.61\linewidth]{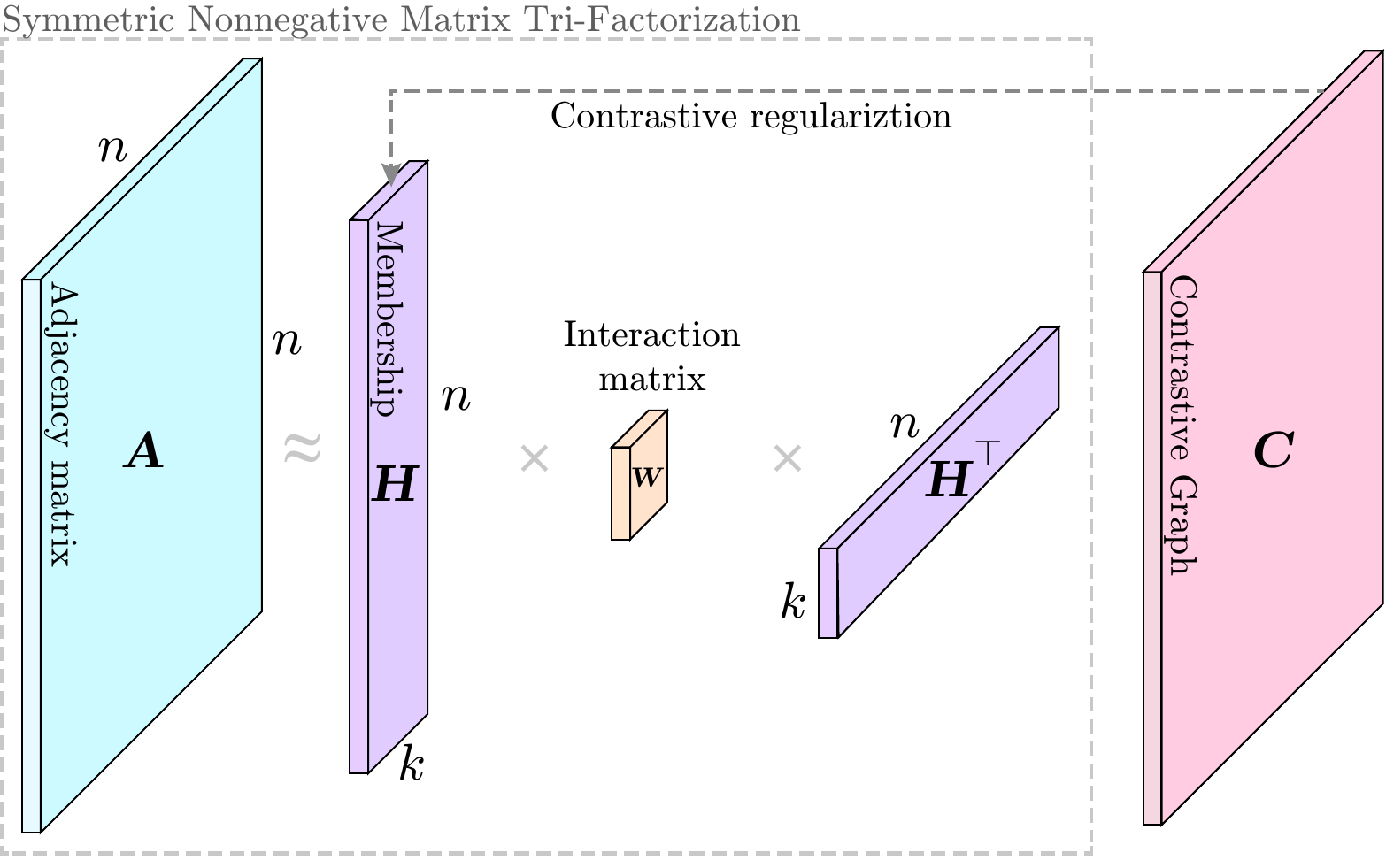}
    \caption{Schematic representation of the iFairNMTF model with contrastive regularization.}
    \label{fig: model block-diagram }
\end{figure}

Symmetric NMF (SNMF) \cite{DBLP:conf/sdm/KuangPD12} is an extension of the traditional NMF that transforms it into a versatile graph clustering model. This model factorizes an adjacency matrix $\bm{A}\in \mathbb{R}_+^{n\times n}$ and is based on the assumption that similar samples ($A_{ij}>0$) should have similar representations ($\bm{h}_i\bm{h}_j^\top>0$) and dissimilar samples ($A_{ij}=0$) should have opposite representations ($\bm{h}_i\bm{h}_j^\top=0$), where $H$ can be interpreted as the \emph{node-to-cluster membership matrix}. More formally:
\begin{align}\label{eq:snmf}
	\min_{\bm{H}\geq 0}\lVert\bm{A}-\bm{HH}^\top\rVert^2_F,
\end{align}
An extended form of the SNMF is the SNM-Tri-Factorization \cite{DBLP:conf/ijcai/PeiCS15} (we omit the "S" and refer NMTF hereafter) which has been tailored to address graph clustering tasks \cite{abdolahi,DBLP:journals/pr/HajiveisehST24}. It takes into account the \emph{cluster-cluster interactions matrix} using an additional factor $\bm{W}$ such that, $A_{ij}\approx\bm{h}^{(i)}\bm{W}{\bm{h}^{(j)}}^\top$. More formally:
\begin{align}\label{eq:snmtf}
	\min_{\bm{H},\bm{W}\geq 0}\lVert\bm{A}-\bm{HWH}^\top\rVert^2_F,
\end{align}
where $\bm{W}$ can be interpreted as the cluster interactions. We build upon this model and extend it into an individual fairness joint learning framework using a contrastive regularization. More formally:
\begin{align}\label{eq:basic}
    \min_{\bm{H},\bm{W}\geq 0}\lVert\bm{A}-\bm{HWH}^\top\rVert^2_F+\lambda\mathcal{R}_{\bm{C}}(\bm{H}),
\end{align} 
Schematically, the model block diagram is illustrated in Figure~\ref{fig: model block-diagram }.
The left term comes from Equation~\eqref{eq:snmtf} and $\mathcal{R}_{\bm{C}}(\bm{H})$ is a contrastive regularization constraining the cluster indicator $\bm{H}$ relatively adjusted by the magnitude of a flexible $\lambda$ parameter ensuring its alignment with group demographics. The contrastive term $\bm{C}=\bm{P}-\bm{N}$ consists of a positive and a negative component:    

\begin{align}\label{eq:graphs}
&
\mathcal{N}_{i,j}=
	\begin{cases}
		1, & \text{if}\; g_i= g_j\\
		0, & \text{otherwise.}
	\end{cases}
&
\mathcal{P}_{i,j}=
	\begin{cases}
		1, & \text{if}\;g_i \neq g_j\\
		0, & \text{otherwise.}
	\end{cases}
 \\\nonumber
 &
 {N}_{ij}=\mathcal{N}_{ij}/\sum_{r=1}^{n}\mathcal{N}_{ir},
 &
 {P}_{ij}=\mathcal{P}_{ij}/\sum_{r=1}^{n}\mathcal{P}_{ir},
\end{align}
which can be enforced to apply the attraction of different demographic groups into the same cluster, and repulsion of same-group members to ensure diversity of their distribution into different clusters according to Equation~\eqref{eq:reg2}:
\begin{align}\label{eq:reg2}
    \min_{\bm{H}}\mathcal{R}_{\bm{C}}=\sum_{i=1}^{n}\sum_{j=1}^{n}\lVert\bm{h}^{(i)}-{\bm{h}^{(j)}}\rVert^2 {C}_{ij}=\mathrm{Tr}(\bm{H}^\top\bm{L}\bm{H}).
\end{align}
where $\bm{{L}}=\bm{{D}}-\bm{{C}}$ is the graph Laplacian and ${D}_{ii}=\sum_{j=1}^n{C}_{ij}$.
By adding the contrastive regularization $\mathcal{R}_{\bm{C}}$ to the NMTF \eqref{eq:snmtf}, we derive the final objective function (loss function) of iFairNMTF, $\mathcal{L}=\mathcal{L}_{\mathcal{F}}+\lambda\mathcal{R}_{\bm{C}}$ as follow:
\begin{align}\label{eq:iFairNMTF}
    \min_{\bm{H},\bm{W}\geq 0}\lVert\bm{A}-\bm{HWH}^\top\rVert^2_F+\lambda\mathrm{Tr}(\bm{H}^\top\bm{L}\bm{H}),
\end{align}

The objective function in Equation~\eqref{eq:iFairNMTF} is a combination, trading-off between the clustering loss and the constrastive regularization term to ensure individual fairness. The hyper-parameter $\lambda \in [0,+\infty)$ controls the compromise between clustering performance and fairness. Smaller $\lambda$ implies a higher importance of the clustering performance and prompts the model to prioritize generating strong and cohesive clusters. Conversely, a higher $\lambda$ prioritizes fairness, prompting the model to create diversified clusters that fairly represent groups of $V_s$.

\subsection{The iFairNMTF Model Optimization}
In this section, we focus on solving the iFairNMTF model.
The objective function in Equation~\eqref{eq:iFairNMTF} is a fourth-order non-convex function with respect to the entries of $\bm{H}$ and has multiple local minima. For these types of problems, it is difficult to find a global minimum; thus a good convergence property we can expect is that every limit point is a stationary point. Therefore, we adopt multiplicative updating rules to update the membership matrix $\bm{H}$ and introduce two Lagrangian multiplier matrices of $\bm{\Theta}$, and $\bm{\Phi}$ to enforce the nonnegative constraints on $\bm{H}$, and $\bm{W}$ respectively, resulting in the following equivalent objective function:
\begin{align}
    \min_{\bm{H},\bm{W}}
    \mathcal{L}
    =\lVert\bm{A}-\bm{HWH}^\top\rVert^2_F+\lambda\mathrm{Tr}(\bm{H}^\top\bm{L}\bm{H})-\mathrm{Tr}(\bm{\Theta}^\top\bm{H})-\mathrm{Tr}(\bm{\Phi}^\top\bm{W}),\nonumber
\end{align}
which can be further rewritten as follows:
\begin{align}
    \min_{\bm{H},\bm{W}}\mathcal{L}
    =&\mathrm{Tr}(\bm{A}^\top\bm{A}-2\bm{A}^\top\bm{HWH}^\top+\bm{HW}^\top\bm{H}^\top\bm{HWH}^\top)\nonumber\\
	&+\lambda\mathrm{Tr} (\bm{H}^\top\bm{L}\bm{H})-\mathrm{Tr}(\bm{\Theta}^\top\bm{H})
 -\mathrm{Tr}(\bm{\Phi}^\top\bm{W}).
\end{align}

The partial derivative of $\mathcal{L}$ with respect to $\bm{H}$ is
\begin{align}
    \frac{\partial\mathcal{L}}{\partial\bm{H}}=&-2\bm{A}^\top\bm{HW}-2\bm{A}\bm{HW}^\top
    +2\bm{HW}^\top\bm{H}^\top\bm{HW}\\\nonumber
    &+2\bm{HW}\bm{H}^\top\bm{HW}^\top
    +2\lambda\bm{L}\bm{H}-\bm{\Theta}.
\end{align}

By setting the partial derivative $\frac{\partial\mathcal{L}}{\partial\bm{H}}$ to 0, we have:
\begin{align}
    \bm{\Theta}=-2\bm{A}^\top\bm{HW}-2\bm{A}\bm{HW}^\top +2\bm{HW}^\top\bm{H}^\top\bm{HW}+2\bm{HW}\bm{H}^\top\bm{HW}^\top
    +2\lambda\bm{L}\bm{H}.
\end{align}

From the Karush-Kuhn-Tucker complementary slackness conditions (KKT), we obtain $\bm{H}\odot\bm{\Theta}=\bm{0}$
where $\odot$ denotes the element-wise product. 
This is the fixed point equation that the solution must satisfy at convergence. By solving this equation, we derive the following updating rule for $\bm{H}$:
\begin{align}\label{eq:updateH}
    \bm{H}\leftarrow\bm{H}\odot\Big(\frac{\bm{A}^\top\bm{HW}+\bm{A}\bm{HW}^\top+\lambda\bm{L}^-\bm{H}}
    {\bm{HW}^\top\bm{H}^\top\bm{HW}+\bm{HW}\bm{H}^\top\bm{HW}^\top
    +\lambda\bm{L}^+\bm{H}}\Big)^{\frac{1}{4}}.
\end{align}
To guarantee the nonnegativity, we separate the positive and negative elements as $\bm{L} = \bm{L}^+ - \bm{L}^-$. Similarly, we differentiate $\mathcal{L}$ with respect to $\bm{W}$ such that:
\begin{align}
    \frac{\partial\mathcal{L}}{\partial\bm{W}}&=-2\bm{H}^\top\bm{AH}+2\bm{H}^\top\bm{HWH}^\top\bm{H}-\bm{\Phi}
\end{align}

By setting the partial derivative $\frac{\partial\mathcal{L}}{\partial\bm{W}}$ to 0, we obtain $\bm{\Phi}$ as:
\begin{align}
    \bm{\Phi}=-2\bm{H}^\top\bm{AH}+2\bm{H}^\top\bm{HWH}^\top\bm{H}.
\end{align}

From the complementary slackness KKT conditions we obtain $\bm{W}\odot\bm{\Phi}=\bm{0}$. 
This is another fixed point equation that the solution must satisfy at convergence. Finally, by solving this equation, we derive the following updating rule for $\bm{W}$:
\begin{align}\label{eq:updateW}
    \bm{W}\leftarrow\bm{W}\odot\frac{\bm{H}^\top\bm{AH}}{\bm{H}^\top\bm{HWH}^\top\bm{H}}.
\end{align}

\begin{algorithm}[tb]
\small
	\caption{\small Individual Fair  Nonnegative Matrix Tri-Factorization (iFairNMTF)}
	\label{alg:alg1}
		\textbf{Input}: adjacency matrix $\bm{A}$, group set $g$, latent factor $k$, trade-off parameter $\lambda$;\\
		\textbf{Output}: cluster assignment $M$; 
	\begin{algorithmic}[1] 
        \STATE Construct the contrastive graph $\bm{C}$ according to \eqref{eq:graphs};
		\WHILE{convergence not reached}
		\STATE Update cluster-membership matrix $\bm{H}$ according to \eqref{eq:updateH};
    \STATE Update cluster-interaction matrix $\bm{W}$ according to \eqref{eq:updateW};
		\ENDWHILE 
        \STATE Calculate cluster assignment $M_i \leftarrow \arg\max (\bm{h}^{(i)}), \forall i \in \{1,{\dots},n\}$
		\STATE \textbf{return} cluster-membership matrix $\bm{H}$ and cluster-interaction matrix $\bm{W}$;
	\end{algorithmic}
\end{algorithm}

%% file: Exp.tex
\subsection{Experimental Setup}
\noindent \textbf{Datasets}
In the paper, six real-world and three synthetic networks are used for benchmarking the performance of the proposed method against competitors. Our synthetic networks are generated according to a generalized Stochastic Block Model (SBM) \cite{DBLP:conf/icml/KleindessnerSAM19} with equal-sized clusters $|C_l|={n}/{k}$ and groups $|V_s|={n}/{g}$ randomly distributed among the clusters. We generate three SBM networks of 2K, 5K, and 10K nodes with k = 5 clusters and g = 5 groups. Real datasets include three high school friendship networks~\cite{mastrandrea2015contact}: \textit{Facebook, Friendship}, and \textit{Contact-Diaries} which represent connections among a group of French high school students. \textit{DrugNet}~\cite{weeks2002social} is a network encoding acquaintanceship between drug users in Hartford, CT. \textit{LastFMNet} \cite{DBLP:conf/cikm/RozemberczkiS20} contains mutual follower relations among users of Last.fm, a recommendation-based online radio and music community in Asia. Lastly, \textit{NBA} is a network containing relationships between around 400 NBA basketball players \cite{DBLP:conf/wsdm/DaiW21}. A detailed description of both real and synthetic datasets, as well as instructions on generating the SBM networks are provided in the supplementary material\footnote{Link to supplemental file and source codes: \href{https://github.com/SiamakGhodsi/iFairNMTF}{Github.com/SiamakGhodsi/iFairNMTF}}. Likewise for dataset statistics including size and number of sensitive groups and also details on cleaning the real datasets.

\noindent \textbf{Competitors}
We compare \textit{iFairNMTF} with four state-of-the-art graph clustering methods, namely, with two group-fair models: i) \textit{Fair-SC} \cite{DBLP:conf/icml/KleindessnerSAM19}, and its scalable version (ii) \textit{sFair-SC}) \cite{wang2023scalable}, iii) an individual-fairness model (\textit{iFair-SC}) \cite{DBLP:conf/nips/GuptaD22} and iv) a deep graph neural network (\textit{DMoN}) \cite{DBLP:journals/jmlr/TsitsulinPPM23}. The three former models are fairness-aware and 
have been already discussed. The latter model is one of the very few DNNs developed for pure graph-partitioning problems, but does not consider fairness. This model extends the general graph neural network (GNN) architecture into a deep modularity optimization GNN. It operates on attributed graphs, thus we pass the sensitive attribute as node-attribute to it. 
The number of layers and learning rate are set according to the official source code provided by the authors (layers = 64 or 512 for small and large networks, $\alpha=0.001$). The number of epochs for DMoN and our method is 500. To produce reliable results, all experiments are averaged over 10 independent runs.

\noindent \textbf{Evaluation measures}
We use accuracy for measuring clustering assignment quality on synthetic networks. For real-world networks, since the ground truth cluster structures are unknown, we use Newman's modularity (Q) measure \cite{doi:10.1073/pnas.0601602103, DBLP:journals/csur/ChakrabortyDMG17} which analyzes the homogeneity of clusters by calculating the proportion of internal links in each cluster for a given partitioning compared to the expected proportion of edges in a null graph with the same degree distribution. Modularity is preferable over cut-based measures due to its robustness against imbalanced cluster sizes. 
We measure the fairness of clustering in terms of the popular average balance ($B$) measure~\cite{wang2023scalable, DBLP:conf/icml/KleindessnerSAM19}:  $B = \frac{1}{k}\sum_{l=1}^{k}Balance(C_l)$, where Balance($C_l$) calculates the minimum group proportion of $C_l$ according to Equation~\eqref{eq:bal}:
\begin{align}\label{eq:bal}
    Balance(C_l)=\min_{s\neq s^{\prime}\in [m]} \frac{\lvert V_s\cap C_l \rvert}{\lvert V_{s^{\prime}}\cap C_l \rvert},
\end{align}
where $l\in [1, k]$ iterates over all the $k$ clusters and $V_s$ identifies each sensitive group of the sensitive attribute.
The minimum balance of each cluster can range between $[0,1]$, thus their average also ranges between $[0,1]$. 

\noindent \textbf{Parameters}
Our model has an adjustable hyper-parameter $\lambda$ to trade-off between the degree of fairness and clustering efficiency (Equation~\eqref{eq:iFairNMTF}). The range of $\lambda$ includes 50 values from $[0,100]$ with a median of $3$ for small and from $[0,3500]$ for large datasets (must be set separately for each dataset.). The effect of $\lambda$ is discussed in Section~\ref{sec: param}. The trade-off parameter $\lambda$ can be set based on user preferences between fairness and clustering quality. A practical way is to select the best value according to the intersection point of $B$ and $Q$, see Figure~\ref{fig:param}.

\subsection{Clustering Quality vs Fairness}
In real datasets, the ground truth partitioning of the networks is unknown, therefore we report the performance for various number of clusters. Figure~\ref{fig:real} illustrates the comparison of our method's results in terms of $Q$ (clustering quality/ modularity) and $B$ (fairness/balance) with those of other models on two datasets, for various numbers of clusters. Dataset balance, highlighted by the yellow dashed line, identifies the proportion of the smallest to the largest group of the sensitive attribute, calculated according to Equation~\eqref{eq:bal}.  
For iFairNMTF, the best $\lambda$ values for each $k$ are used. They are selected based on the intersection of $Q$ and $B$ charts as in Figure~\ref{fig:param}: $\lambda=2$ for DrugNet, and $\lambda=100$ for LastFM.
\input{fig_results}
As we can see from this figure, our model outperforms the SC-based models in terms of both measures on both datasets. It reports a lower clustering quality $Q$ on LastFM than DMoN which is a neural model primarily focusing on identifying the most modular partitioning of the graph through modularity optimization. DMoN's $Q$ outcomes reveal varied patterns on LastFM and DrugNet, attributed to differences in network size and density. Neural models typically excel in data-intensive learning cycles, yielding better performance on larger datasets. For instance, LastFM, a substantially larger graph with 5k nodes and 20k edges, showcases this advantage compared to the 200-node DrugNet.
However, in terms of fairness, DMoN fails to generate diverse clusters w.r.t. the sensitive attribute, as evidenced by low balance ($B$). In contrast, our model consistently achieves well-distributed clusters, boasting the highest balance scores among all competitors.

Next, we compare all the models, on all the datasets with a fixed number of clusters $k=5$, the median of our selected number of clusters. 
The results are presented in Table~\ref{table:results}. 
We distinguish between real and SBM networks based on the measurable accuracy of partitioning quality in SBM networks, where ground-truth clusters are known. Additionally, we present the average modularity ($Q$) and balance ($B$) across all clusters for real datasets.
\input{table_results}

The results on real networks indicate the superiority of our proposed iFairNMTF model while reporting the best $Q$ values on 5/6 (meaning 5 out of 6) datasets and 3/6 w.r.t. $B$. Similarly, on SBM networks iFairNMTF stands the best with 2/3 best accuracy and balance scores. It is worth noting that, in the SBM experiment, DMoN and sFairSC failed to deliver the required number of clusters resulting in empty clusters implying inconsistency in accuracy calculation since the cluster assignments need to be masked to be comparable to true labels. 

\subsection{Parameter Analysis} \label{sec: param}
This section studies the effect of the $\lambda$ hyper-parameter on the iFairNMTF model’s performance in terms of $Q$ and $B$ for $k = 5$ clusters in comparison with the performance of other models. 
In this experiment, we also provide the results of the vanilla SC and vanilla NMTF (the same as iFairNMTF with $\lambda=0$) models. 
The results are illustrated in Figure~\ref{fig:param}.
Based on the results, a comparably good value for the $\lambda$ parameter can be selected in the intersection of the two measures. These twin charts provide a nice opportunity to visualize the distribution of results and make it easy to select. For instance, values in the range $[0.1, 4]$ for Drugnet and $[55, 200]$ for LastFM are suggested. It gives the end-user a desirable autonomy and depends on the user's demands on how to select values for this parameter. Consider that, since LastFM is a much larger network than Drugnet, we increase the range of $\lambda$ with 100 values from $0$ to $\lambda=5000$. Complementary results can be found in the supplementary material (see footnote 4).
\input{fig_param}

\subsection{Interpretability Analysis}
\begin{figure}
    \centering    
    \captionsetup{font=scriptsize} 
    {\includegraphics[width=0.35\linewidth]{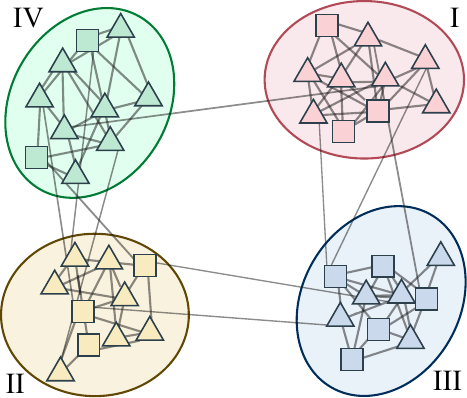}} \quad\quad %
    {\includegraphics[width=0.3\linewidth]{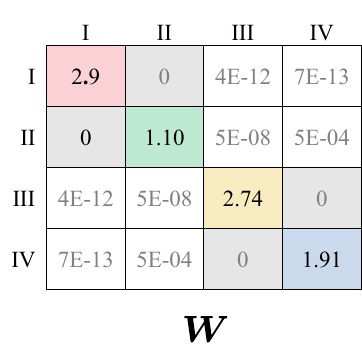}}%
    \caption{Interpretability of $\bm{W}$ factor for a 40-node graph divided to 4 clusters. Shapes indicate groups.}
    \label{fig:interpret}
\end{figure}

In this section, in addition to the inherent model interpretability through the direct clustering given by the $\bm{H}$ factor, we investigate the explicitly interpretable intermediary factor $\bm{W} \in \mathbb{R}^{k\times k}_{+}$ of the iFairNMTF model introduced in Equation~\eqref{eq:snmtf}. This factor is a symmetric square matrix consisting of non-negative scores representing the strength of cluster-cluster interactions. Diagonal elements reflect intra-cluster connectivity such that the score for dense clusters is expected to be higher. An illustrative example of a graph with 40 nodes distributed between 4 clusters and an imbalanced group distribution of $35\%$ (square shape) to $65\%$ (triangle shape) is shown in Figure~\ref{fig:interpret}. We apply Algorithm~\ref{alg:alg1} to this graph with $\lambda=1$, and the model identifies the true clusters. Entries corresponding to clusters like \RomanNumeralCaps{1}$-$\RomanNumeralCaps{2}, which have no interactions (no links) together, are assigned a value of $0$. Furthermore, the score for clusters \RomanNumeralCaps{4}$-$\RomanNumeralCaps{1} is notably lower compared to \RomanNumeralCaps{4}$-$\RomanNumeralCaps{2}, reflecting the difference in the number of connecting links between these clusters.   

%% file: fig_results.tex
\begin{figure}
    \centering    
    \captionsetup{font=scriptsize} 
    {\includegraphics[width=0.495\linewidth]{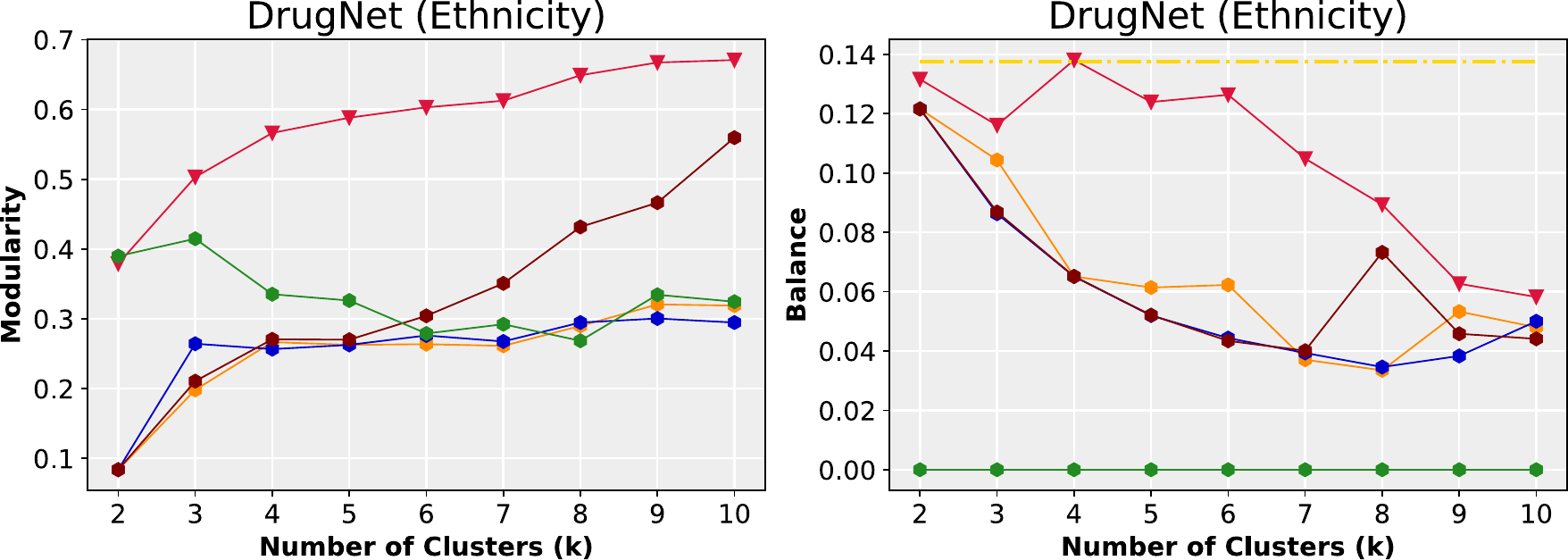}} \hfill %
    {\includegraphics[width=0.495\linewidth]{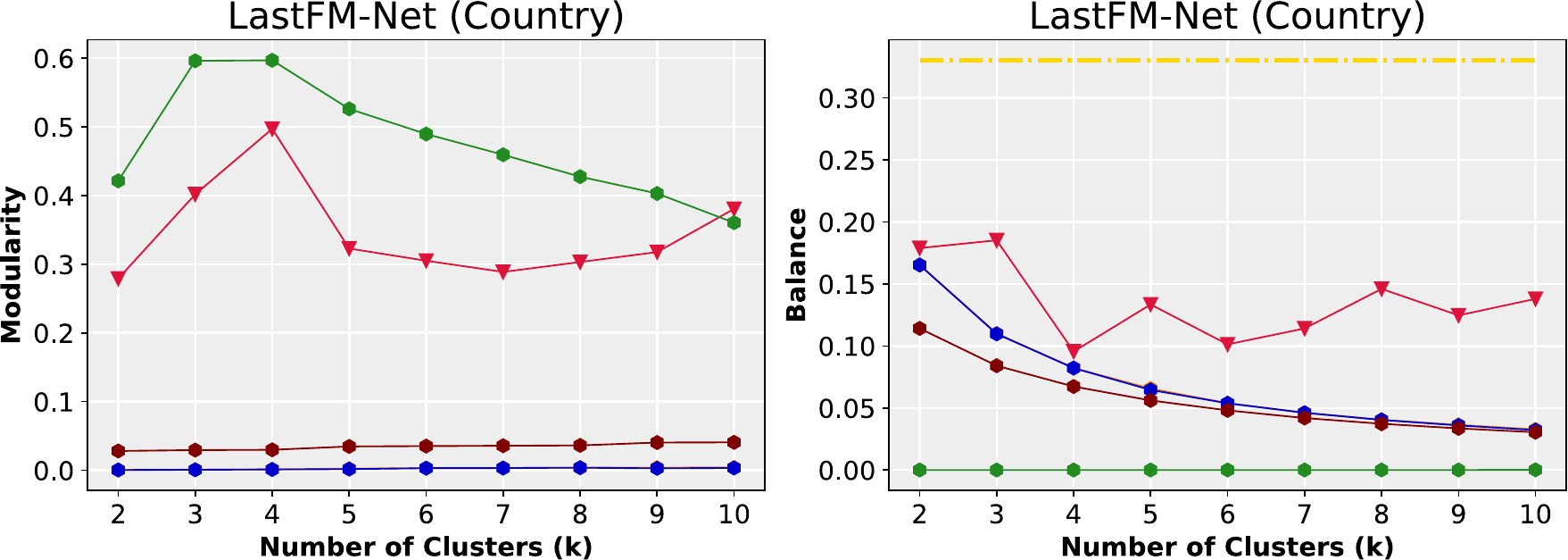}}%

    {\includegraphics[width=\linewidth]{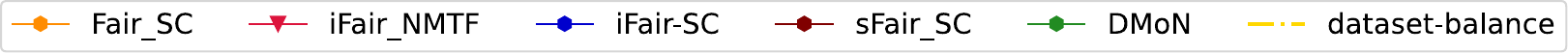}}
    \caption{Performance comparison w.r.t. clustering quality/modularity $Q$ and cluster fairness $B$ (higher values are better for both measures) on DrugNet, and LastFM for different number of clusters $k\in [2,10]$. $k=10$ is the convergence point of all models. }
    \label{fig:real}
\end{figure}

%% file: table_results.tex
\begin{table}
    \captionsetup{font=scriptsize}
    \scriptsize
    \centering
    \setlength{\tabcolsep}{4.3pt}
    \caption {\label{table:results} Results illustrating modularity (Q) and average balance (B) of real networks, and accuracy (Acc) and average balance (B) results on SBM networks for $k=5$ clusters.(\underline{\textbf{Bold-underline}}) and \underline{underline} indicate best and second best B results. Best Q, Acc are highlighted with \colorbox{lightgray!60}{\textbf{boldfaced gray}}. }
    
\begin{tabularx}{\linewidth}{lcccccccccc}
    \toprule
    \multirow{2}{*}{\bf{Network}}&  \multicolumn{2}{c}{\bf{FairSC}} & \multicolumn{2}{c}{\bf{sFairSC}} & \multicolumn{2}{c}{\bf{iFairSC}} & \multicolumn{2}{c}{\makecell{\bf{DMoN}}} & \multicolumn{2}{c}{\makecell{\bf{iFairNMTF}}}\\
    
    \cmidrule(lr){2-3}
    \cmidrule(lr){4-5}
    \cmidrule(lr){6-7}
    \cmidrule(lr){8-9}
    \cmidrule(lr){10-11}

     & 
    \bf{B} & 
    \bf{Q} & 
    \bf{B} & 
    \bf{Q} &
    \bf{B} & 
    \bf{Q} & 
    \bf{B} & 
    \bf{Q} &
    \bf{B} & 
    \bf{Q} \\

    \midrule

    Diaries & \underline{0.708} & 0.612 & \textbf{\underline{0.809}} & \textbf{\colorbox{lightgray!60}{0.684}} & 0.699 & 0.647 & 0.263 & 0.145 & 0.648 & 0.640 \\
    
    Facebook & 0.327 & 0.449 & \textbf{\underline{0.602}} & 0.500 & 0.330 & 0.448 & 0.268 & 0.048 & \underline{0.514} & \textbf{\colorbox{lightgray!60}{0.509}} \\
    
    Friendship & 0.391 & 0.483 & \underline{0.485} & 0.627 & 0.374 & 0.392 & 0.183 & 0.140 & \textbf{\underline{0.631}} & \textbf{\colorbox{lightgray!60}{0.669}} \\
    
    DrugNet & 0.052 & 0.263 & 0.052 & 0.270 & \underline{0.061} & 0.263 & 0.000 & 0.326 & \textbf{\underline{0.124}} & \textbf{\colorbox{lightgray!60}{0.588}} \\
    
    NBA & 0.083 & 0.000 & \textbf{\underline{0.323}} & 0.113 & 0.072 & 0.000 & 0.036 & 0.057 & \underline{0.286} & \textbf{\colorbox{lightgray!60}{0.150}} \\
    
    LastFM & 0.065 & 0.003 & 0.056 & 0.035 & \underline{0.066} & 0.002 & 0.000 & 0.526 & \textbf{\underline{0.069}} & \textbf{\colorbox{lightgray!60}{0.600}} \\

    \midrule

    & 
    \bf{B} & 
    \bf{Acc} & 
    \bf{B} & 
    \bf{Acc} &
    \bf{B} & 
    \bf{Acc} & 
    \bf{B} & 
    \bf{Acc} &
    \bf{B} & 
    \bf{Acc} \\
    
    \hdashline

     SBM-2K & \underline{0.575} & 0.588 & -- & -- & 0 & 0.799 & -- & -- & \textbf{\underline{0.953}} & \textbf{\colorbox{lightgray!60}{0.958}} \\

     SBM-5K & \textbf{\underline{0.995}} & \textbf{\colorbox{lightgray!60}{0.998}} & -- & -- & 0 & 0.799 & -- & -- & \underline{0.941} & 0.962 \\
     
     SBM-10K & \underline{0.999} & 0.999 & -- & -- & 0 & 0.600 & -- & -- & \bf{\underline{1}} & \textbf{\colorbox{lightgray!60}{1}} \\
    
    \bottomrule
\end{tabularx}
\end{table}

%% file: fig_param.tex
\begin{figure*}
    \captionsetup{font=scriptsize}
    \centering
    {\includegraphics[width=0.495\linewidth]{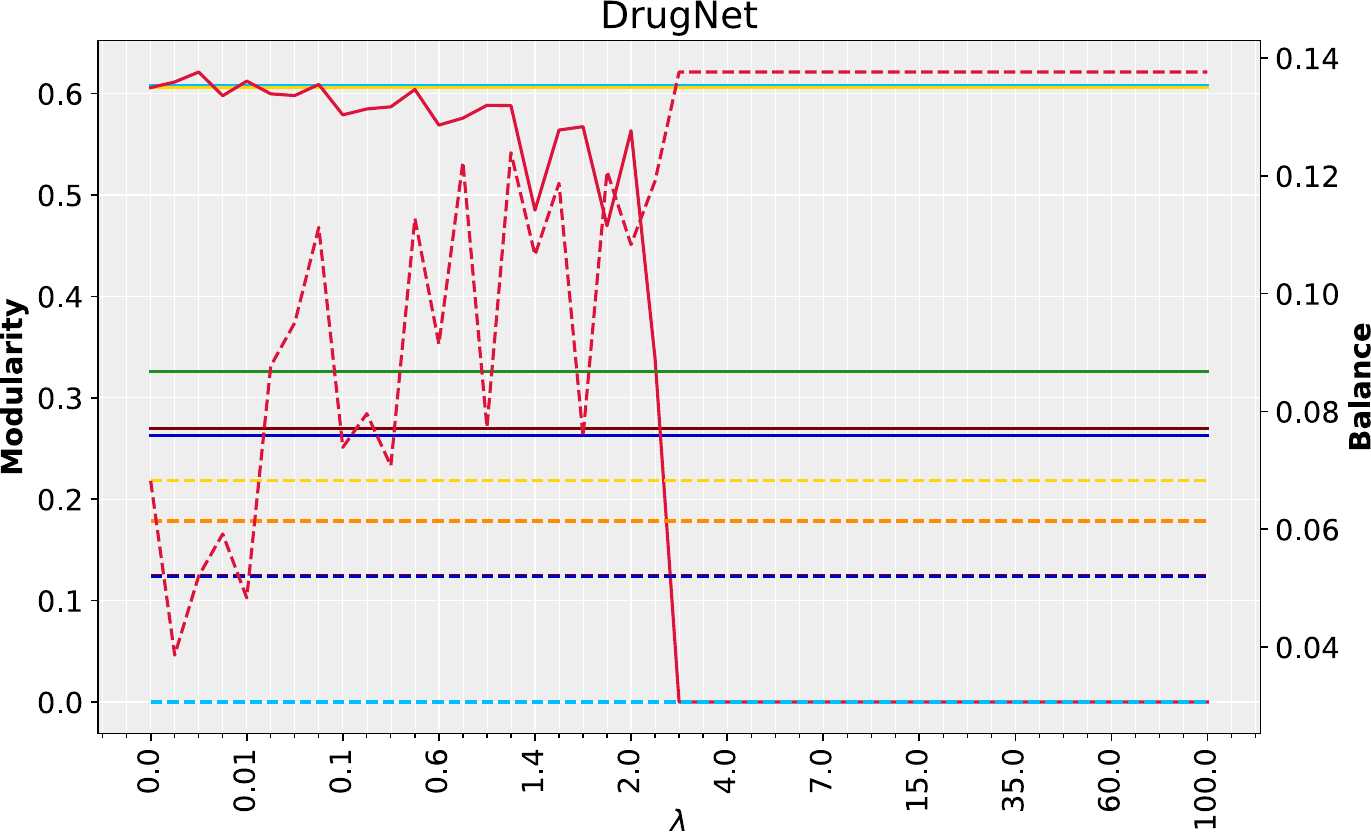}}\hfill  {\includegraphics[width=0.495\textwidth]{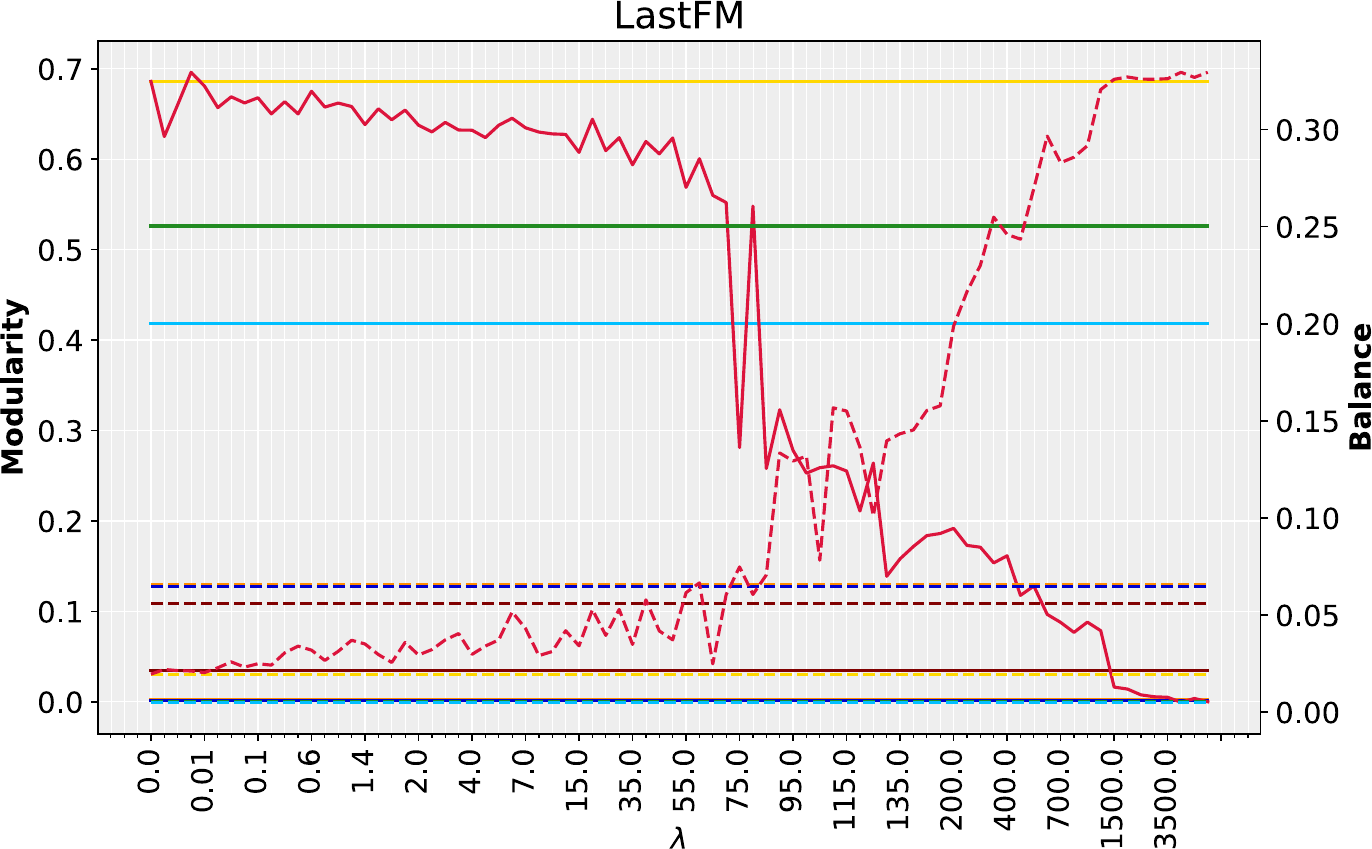}}
    {\includegraphics[width=\textwidth]{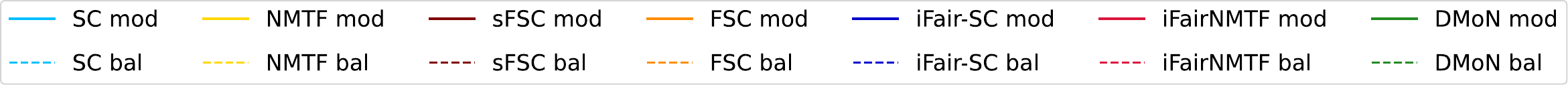}}
    \caption{Parameter $\lambda$ analysis of the iFairNMTF on Drugnet and LastFM-Net datasets with k = 5 in terms of Q and B for $\lambda\in [0,100]$. Solid lines depict modularity and dashed lines represent balance. Only the behavior of FairSNMF depends on $\lambda$.}
    \label{fig:param}
\end{figure*}

%% file: Acknowledgement.tex
\subsubsection{Acknowledgements}
This work has received funding from the European Union’s Horizon 2020 research and innovation programme under Marie Sklodowska-Curie Actions (grant agreement number 860630) for the project ‘’NoBIAS - Artificial Intelligence without Bias’’. This work reflects only the authors’ views and the European Research Executive Agency (REA) is not responsible for any use that may be made of the information it contains. The research was also supported by the EU Horizon Europe project MAMMOth (GrantAgreement 101070285).